\documentclass[letterpaper, 10 pt, conference]{ieeeconf}  

\IEEEoverridecommandlockouts                              

\usepackage{graphicx} 
\usepackage{epsfig} 
\usepackage{times} 
\usepackage{amsmath} 
\usepackage{amssymb}  
\usepackage[ruled,vlined,linesnumbered]{algorithm2e}
\usepackage{booktabs}
\usepackage{multirow}
\usepackage{subcaption}
\usepackage{cite}
\usepackage{hyperref}

\title{\LARGE \bf
Navigating in Uncertain Environments with Heterogeneous Visibility
}

\author{Jongann Lee$^{1}$ and Melkior Ornik$^{1}$
\thanks{$^{1}$ Jongann Lee and Melkior Ornik are with the University of Illinois
Urbana-Champaign (UIUC), Urbana, IL 61801, USA.
        {\tt\small jongann2, mornik@illinois.edu}}%
}

\begin{document}

\maketitle
\thispagestyle{empty}
\pagestyle{empty}

\begin{abstract}

Navigating an environment with uncertain connectivity requires a strategic balance between minimizing the cost of traversal and seeking information to resolve map ambiguities. Unlike previous approaches that rely on local sensing, we utilize a framework where nodes possess varying visibility levels, allowing for observation of distant edges from certain vantage points. We propose a novel heuristic algorithm that balances the cost of detouring to high-visibility locations against the gain in information by optimizing the sum of a custom observation reward and the cost of traversal. We introduce a technique to sample the shortest path on numerous realizations of the environment, which we use to define an edge's utility for observation and to quickly estimate the path with the highest reward. Our approach can be easily adapted to a variety of scenarios by tuning a single hyperparameter that determines the importance of observation. We test our method on a variety of uncertain navigation tasks, including a map based on real-world topographical data. The method demonstrates lower mean cost of traversal compared to a shortest path baseline that does not consider observation and has exponentially lower computational overhead compared to an existing method for balancing observation with path cost minimization. Code is available at \url{https://github.com/jongann-lee/CTP_with_Heterogeneous_Visibility/tree/submission1}.

\end{abstract}


\section{INTRODUCTION} \label{sec:Introduction}

Navigating uncertain environments is a classical problem in robotics ~\cite{papadimitriou1991CTP,bnaya2009CTP_remote_sensing,wilde2022online, yu2025multi}. A robot on mountainous terrain can utilize satellite and topography maps to find the easiest path to a target. However, random events such as mud pools from rain, or knocked down trees can block previously traversable paths. Because these uncertainties are often revealed only through observation, an optimistic planner that assumes all edges are clear may commit to paths that end up leading to significant backtracking. These uncertainties can be resolved by traversing to a vantage point, such as the top of a hill. Observing obstacles early can reduce the amount of backtracking, but visiting a high-visibility location to do so often incurs an additional cost \cite{muguira2023visibility}. Therefore, balancing the cost of reaching a vantage point with the benefit of better observation is crucial for effective navigation in uncertain environments.

Existing frameworks for navigating under uncertainty often overlook the strategic utility of high-visibility locations. Most previous methods assume all locations have identical visibility ~\cite{papadimitriou1991CTP,wilde2022online,yu2025multi}. Other works consider high-visibility locations, but prioritize maximizing information gain rather than minimizing the total traversal cost ~\cite{lim2016RAId,vashisth2024RL_IPP}.



We propose a path planning framework that balances the benefit of reaching high-visibility locations against the additional traversal cost. We define an observation reward that quantifies the utility of observation and add it to the cost of traversal to form the total path reward. Notably, our method does not require an explicit model of edge blockage location or probability. In summary, our contributions are as follows:
\begin{enumerate}
    \item We introduce a problem formulation for navigation under uncertain connectivity where visibility is location-dependent and observation is passive and cost-free, distinguishing it from CTP with remote sensing and from IPP.
    \item We define an edge utility based on path frequency across sampled blockage realizations, and use it to construct an observation reward that is balanced against traversal cost through a single tunable parameter $\lambda$, with the shortest-path planner recovered at $\lambda = 0$.
    \item We evaluate our method on synthetic, procedurally generated, and real-terrain environments, demonstrating lower expected traversal cost than the shortest-path baseline and exponentially lower runtime than an existing observation-aware planner ~\cite{macdonald2019RPP}.
\end{enumerate}

\section{PROBLEM FORMULATION} \label{sec:Problem_Formulation}

We now formally define the problem of navigating in uncertain environments with heterogeneous visibility. The environment is modeled as a directed graph $G = (\mathcal V, \mathcal E)$ where nodes $v_i \in \mathcal V$ and edges $e_{ij} = (v_i , v_j) \in \mathcal E$ represent discrete locations and traversal paths, respectively. A traversal cost $c(e) \in \mathbb R^+$ is defined for each edge and is known \textit{a priori}. A visibility function $\mathrm{vis}(v_i) \subseteq \mathcal E$ is defined as the set of all visible edges from the node $v_i$. The visibility function is assumed to be provided alongside the environment map.

Although $G$ is directed, the connectivity is symmetric: for any pair of nodes, either both directed edges exist or neither does. However, the cost of traversal $c(e)$ may differ between the two, such as the varying energy cost of uphill versus downhill. 

We denote a path on the graph $G$ as a sequence of adjacent nodes $P = [v_1, v_2, \cdots, v_n]$, where $v_i\in \mathcal V$ and each consecutive pair $(v_i, v_{i+1}) \in \mathcal E$. Let $e \in \mathcal E_P$ denote an edge $e = (v_i, v_{i+1})$ connecting consecutive nodes in the sequence. 

The environment $G$ contains blocked edges. These blockages are bidirectional and static; an edge's status remains constant for the duration of the mission. We assume no prior knowledge on the probability distribution of edge blockage. 

The objective is to navigate from the given source node $v_s \in \mathcal V$ to the given target node $v_t \in \mathcal V$ while minimizing the total traversal cost $\sum_{e \in\mathcal E_P} c(e)$. 

\section{RELATED WORKS}

The shortest path problem on a graph with uncertain edges is commonly referred to as the \textit{Canadian Traveler Problem} (CTP) \cite{papadimitriou1991CTP}. Given a graph $G = (\mathcal V, \mathcal E)$ with source node $v_s \in \mathcal V$, target node $v_t \in \mathcal V$, and traversal cost $c(e)$, the objective is to navigate from $v_s$ to $v_t$ such that the total traversal cost $\sum c(e)$ is minimized. Each edge is blocked with probability $p(e)$, and its status is observed when the agent reaches an adjacent node. The blockages are stochastically determined prior to the agent's first action and do not change subsequently. CTP has been shown to be $\#$P-hard and PSPACE-complete, making it difficult to solve analytically \cite{fried2013CTP_complexity}. Thus, a wide range of heuristic methods have been proposed, including approximation methods \cite{lim2017shortest}, sampling methods \cite{eyerich2010high}, and graph search algorithms with heuristic pruning to reduce the search space \cite{aksakalli2016based}. 

Numerous variants of the CTP have been proposed and studied. LAMP assumes correlations between edge blockage probabilities, learning an estimate over successive iterations \cite{tsang2021lamp}. Guo et al. optimize the linear combination of the mean and the standard deviation of the path's cost \cite{guo2022dual}. There has also been research on the multi-agent, multiple-target variant of CTP \cite{wilde2022online,liu2021mSTSP}. However, all of these variants do not model differences in visibility; nodes can only observe adjacent edges. 

There is a variant of CTP that models differences in visibility: \textit{CTP with remote sensing}, where the agent can observe any edge $e$ from any node $v$ with an observation cost $C_{obs}(v,e)$ \cite{bnaya2009CTP_remote_sensing}. The proposed solution is to calculate the value of information (VOI): the expected decrease in the traversal time by observing the edge. If VOI is larger than the cost of observation, then the agent observes the edge. However, our problem formulation differs in two ways: we assume that only a subset of edges are visible from a given node, and that observation is a passive process with zero additional cost. Consequently, VOI is not directly applicable to our problem.

Another closely related problem is \textit{Informative Path Planning} (IPP), which is defined on the same graph with uncertain edges as CTP but maximizes the information gain with a constraint on the total traversal cost. Classical IPP approaches compute the trajectory offline using computationally intensive methods from combinatorics, such as branch-and-bound \cite{binney2012branch}\cite{arora2017RAOr}. More recently, researchers have proposed frameworks capable of online replanning to incorporate newly acquired information during execution, leveraging adaptive heuristics and reinforcement learning \cite{lim2016RAId}\cite{vashisth2024RL_IPP}. However, IPP optimizes for observation under cost constraints, representing an inversion of our problem's priorities. Because our objective is to minimize traversal cost while using observation as a secondary incentive for robustness, traditional IPP methods are not directly applicable.

The \textit{Reactive Planning Problem} (RPP) proposed by MacDonald et al. works with a problem most similar to ours \cite{macdonald2019RPP}. In the RPP framework, the environment is modeled as a graph $G$, where a set of possible graph configurations $G_i$ and their associated probability distribution $p(G_i)$ are assumed to be known \textit{a priori}. Different nodes have different visibility with a cost $\mu(O)$ for performing an observation $O$. The objective is to minimize the expected cost of traversal over the distribution of possible configurations. This is achieved by comparing two options: going to the goal using only the known edges, and going to a high-visibility location first then to the goal. The two options are scored using a sum of expected traversal cost and reduction in uncertainty. While this approach demonstrates good performance on various uncertain graph environments, its explicit use of the entire configuration space limits its scalability, as the number of possible scenarios increases exponentially with the number of potential edge blockages.

\section{METHOD} \label{sec:method}

We now present our method for solving the problem in Section \ref{sec:Problem_Formulation}. Our solution for generating paths that balance observation and path cost minimization is to maximize the reward 
\begin{equation} \label{eq:reward}
    R(P) = \lambda R_{obs}(P) - \sum_{e \in \mathcal E_P} c(e),
\end{equation}
where $R_{obs}(P)$ is the observation reward, which we include to incentivize choosing the path with higher visibility. The observation reward of node $v_i \in P$ is defined as the sum of all the visible, previously unobserved edges' utility $U(e)$. We define $U(e)$ to quantify the importance of an edge to the overall objective.

The parameter $\lambda$ is a hyperparameter that is tuned by the user using the estimated probability of edge blockage and the overall graph structure. If the edges are more likely to be blocked and the cost of detouring from a blocked edge is high, then it is better to increase $\lambda$. 

This section is organized as follows. First, we introduce a method for sampling a set of short, diverse paths, which we utilize for estimating the edge value as well as heuristically determining the path with the highest overall reward. Next, we define the edge utility and the observation reward. Finally, we present the overall method for determining the path that maximizes $R(P)$.

\subsection{Short Diverse Path Sampling} \label{subsec:sampling}

\begin{figure}[t]
    \centering
    \includegraphics[width=\columnwidth]{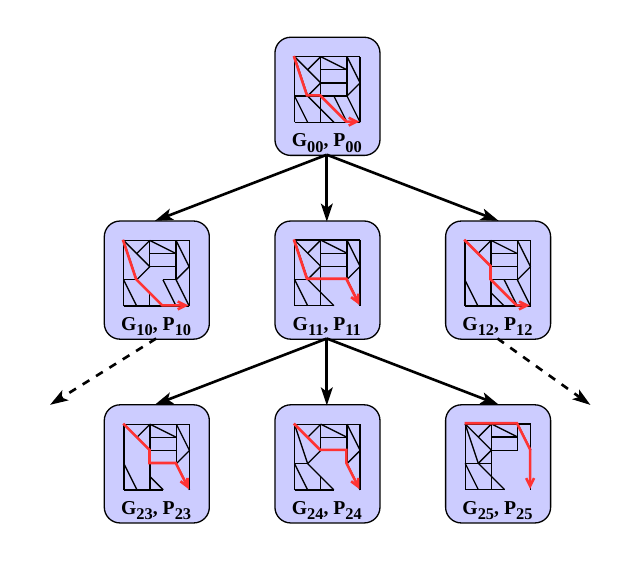}
    \caption{\textbf{Short Diverse Path Sampling:} We generate a set of short diverse paths by repeatedly generating the shortest path, then placing an obstacle on the path to generate a blocked environment. The generated set of paths forms a rooted tree.}
    \label{fig:sampling_tree}
    \vspace{-0.1in}
\end{figure}

On a graph $G = (\mathcal V, \mathcal E)$ with $v_s, v_t$, and edge blockages $\mathcal E_b$, the shortest path on $G' = (\mathcal V, \mathcal E \backslash \mathcal E_b)$ will be short, but distinct from the shortest path on $G$. Therefore, a set of shortest paths on numerous blocked realizations of $G$ forms a set of \textit{short, diverse} paths. We use this set of paths to estimate the utility of an edge and as a search domain for maximizing the overall reward $R(P)$. This subsection illustrates how we sample such a set.

To sample a set of short diverse paths, we utilize an approach inspired by Voss et al. \cite{voss2015short_diverse_path}. Given a graph $G$ with edge cost $c(e)$, source node $v_s$, and target node $v_t$, we compute the initial shortest path using Dijkstra's algorithm \cite{cormen2022intro_algorithms}. This path, along with the original graph, constitutes the root node of a rooted tree. We generate child nodes by randomly sampling $n$ nodes along the current shortest path to serve as centers for virtual obstacles. Each virtual obstacle is defined as the set of nodes within an $m$-hop radius of a sampled center and all edges that connect two virtual obstacle nodes are removed. For each of the $n$ modified graphs, we compute a new shortest path which forms the children of the current tree node. We continue this recursive branching until the tree reaches a specified depth $r$. The process yields a maximum of $\sum_{i=0}^r n^i$ nodes, each representing a shortest path on different blockage realizations. If the addition of an obstacle disconnects $v_s$ from $v_t$, no path is recorded for that specific node. Unlike the original approach, we maintain an explicit tree structure and avoid similarity based pruning; this preserves path frequency across different blockage realizations as a measure of importance.

Sampling paths using a tree has two benefits compared to randomly placing obstacles. First, randomly placed obstacles are not going to change the shortest path if the original shortest path is not obstructed. Our sampling method always places obstacles on the shortest path at that particular configuration, ensuring that all obstacles alter the shortest path. Second, the depth of the tree yields an order of paths. Paths at depth $i$ are only going to be the shortest path when all $i$ specific obstacles are present. Thus, paths at higher depth are less likely to be the shortest path. 

\subsection{Edge Utility and Observation Reward} \label{subsec:observation_reward}

To incentivize the agent to visit high-visibility locations, we define an observation reward based on the number of newly discovered edges. However, not all edges of the graph are equally useful; some are located far away from either $v_s$ or $v_t$ making their status irrelevant to the objective. Therefore, we define the utility of an edge to quantify its relevance to the overall objective.

The utility of each edge $U(e), e \in \mathcal E$ is initialized to zero. A set of diverse short paths $\mathcal P_0$ is sampled using the method outlined in subsection \ref{subsec:sampling}. Let $P_{ij} \in \mathcal P_0$ denote the $j$th path at depth $i$ of the tree of sampled paths. The utility of each edge is defined as:
\begin{equation} \label{eq:edge_utility}
    U(e) = \sum_{P_{ij} \in \mathcal P_0:e \in P_{ij}} \frac{1}{n^i (r+1)},
\end{equation}
where $n$ is the number of virtual obstacles per path and $r$ is the height of the tree. This definition ensures that utility is normalized such that $0 \leq U(e) \leq 1$. 

The observation reward for a node $v_i \in P$ is then defined as the sum of utilities for all the visible, unobserved edges

\begin{equation} \label{eq:observation_reward}
    R_{obs}(v_i | v_{i-1}, \cdots , v_1) = \sum_{e \in \mathrm{vis}(v_i)} U(e) \left( 1 - \mathbf 1_{obs}(e) \right).
\end{equation}
The total path observation reward is then
\begin{equation}
    R_{obs}(P)= \sum_{v_i \in P} R_{obs}(v_i | v_{i-1}, \cdots , v_1).
\end{equation}

To prohibit the abuse of the observation reward by repeatedly observing the same edge, we incorporate a \textit{has been observed} indicator function $\mathbf 1_{obs}$ which is \textit{true} if the edge has been previously observed and \textit{false} otherwise.

From the definition in \eqref{eq:observation_reward}, $R_{obs}$ depends on the previous path taken by the agent. Therefore, it is not Markovian. 

\subsection{Path Reward Maximization} \label{subsec:our_method}

\begin{algorithm}[t]
    \caption{Online Path Reward Maximization} \label{alg:main_method}

    \DontPrintSemicolon
    \SetKwInOut{Input}{Input}
    \SetKwInOut{Output}{Output}

    \Input{Graph (previously known map) $G = (\mathcal V, \mathcal E)$ with edge traversal cost $c(e) \in \mathbb R^+$, visibility mapping $\mathrm{vis}(v) \subset \mathcal E$, source node $v_s \in \mathcal V$ and target node $v_t \in \mathcal V$}

    \BlankLine
    $U(e) \gets 0, \; \mathbf 1_{obs}(e) = \mathrm{False}, \quad \forall e \in \mathcal E$ \;
    $\mathcal P_0 \gets \mathrm{ShortDiversePaths}(G, v_s, v_t)$\;
    $U(e) \gets U(e) + \frac{1}{n^i(r+1)} \quad \forall e \in P_{ij}, \; \forall P_{ij} \in \mathcal P$ \;
    $v_{curr} \gets v_s$ \;
    $\mathcal P \gets \mathrm{ShortDiversePaths}(G, v_{curr}, v_t)$ \;
    $P \gets \arg \max_{P \in \mathcal P} R(P) $ \;
    
    \While{$ v_{curr} \neq v_t$}{
        $ \mathcal E_{visible} \gets \mathrm{vis}(v_{curr})$ \;
        $\mathbf 1_{obs}(e) = True, \quad \forall e \in \mathcal E_{visible}$\;
        \If{ $\exists e \in \mathcal E_{\text{visible}}$ such that $e$ is blocked }{
            $\mathcal{E} \gets \mathcal{E} \setminus \{e \in \mathcal E_{\text{visible}} \mid e \text{ is blocked}\}$ \;
            \If{ any $e \in P$ is blocked } {
                $\mathcal{P} \gets \mathrm{ShortDiversePaths}(G, v_{\text{curr}}, v_t)$ \;
                $P \gets \arg \max_{P \in \mathcal{P}} R(P)$ \;
            }
            
        }
        $ v_{curr} \gets \text{Next node in } P$ \;
    }
\end{algorithm}

The overall framework of our method is a path reward maximization scheme conducted via sampling. The total path reward is defined as in \eqref{eq:reward} with the observation reward defined in the previous section.

Since the reward is non-Markovian, the majority of heuristics for path reward optimization are not applicable as they are variations of dynamic programming and thus require the reward to be Markovian \cite{fu2006shortest_path_review}. Given these constraints, we employ a Monte Carlo sampling approach to identify the path that maximizes the total reward. Using the technique detailed in Section \ref{subsec:sampling}, we generate a set of short, diverse candidate paths $\mathcal P$. The reward for each candidate path is then evaluated, and the path yielding the highest total reward $P = \arg\max_{P \in \mathcal P} R(P)$ is selected for execution.

An overview of the proposed method is presented in Algorithm \ref{alg:main_method}. Given the environment graph $G$, we first calculate the edge utility $U(e)$ for all edges $e \in \mathcal E$ by sampling a set of short, diverse paths $\mathcal P_0$. An initial best path is then determined by generating a secondary set of candidate paths $\mathcal P$ and selecting the path that maximizes the total path reward $R(P)$. The agent traverses along this path and updates its internal map of the environment, including the \textit{has observed} indicator variable $ \mathbf 1_{obs}$. If the agent observes that an edge on the intended path is blocked, it recalculates the best path using the updated map. 

\section{EXPERIMENTS} \label{sec:experiments}

In this section we empirically demonstrate the utility and flexibility of our method on a variety of uncertain navigation tasks. First, we analyze the performance on a simple plateau environment that consists of high-visibility plateaus and narrow corridors in between the plateaus that can be blocked. Next, we procedurally generate numerous plateau maps, showcasing that our algorithm can be adapted to a wide variety of environments. Finally, we test our method on a natural terrain map generated using OpenTopography, demonstrating that our method is useful in real-world environments \cite{OpenTopography}. 

We compare three algorithms:
\begin{enumerate}
    \item \textbf{Shortest Path (SP):} A baseline that follows the minimum cost path and replans only upon observing a blockage on the intended path \cite{bnaya2009CTP_remote_sensing}.
    \item \textbf{Reactive Planning Problem (RPP):} The method proposed by MacDonald et al. \cite{macdonald2019RPP}.
    \item \textbf{Our Method:} Evaluated at the pre-determined optimal reward weight.
\end{enumerate}
Shortest path is the baseline algorithm that does not take visibility into account and RPP is the comparable existing method that utilizes visibility when planning in uncertain environments. 

One point to note is that our method's runtime does not scale with the number of possible blocked graph realizations. On the other hand, RPP scales exponentially at $2^N$ with the number of possible blockage locations $N$ since it calculates the shortest path for all blocked graph realizations. Experiments in subsection \ref{subsec:procedurally_gen_env} block up to 24 edges for which RPP is estimated to require 100 hours per run. Hence, RPP is only tested on the single plateau environment in Subsection \ref{subsec:plateau_experiments}.

When deploying our method it is important to tune the reward weight $\lambda$. The optimal value of $\lambda$ depends on the probability of edge blockage and structure of the graph $G$. The effect of $\lambda$ on our method's performance will be thoroughly analyzed throughout this section, including an ablation study in subsection \ref{subsec:ablation}. 

Our method is a generalization of SP since it simplifies into the SP baseline when $\lambda = 0$. Therefore, the purpose of the comparison with SP is to showcase that our method can adapt to a wider range of uncertain scenarios better than the SP baseline.

\subsection{Plateau Environment} \label{subsec:plateau_experiments}

\begin{figure}[t]
    \centering
    \includegraphics[width=\columnwidth]{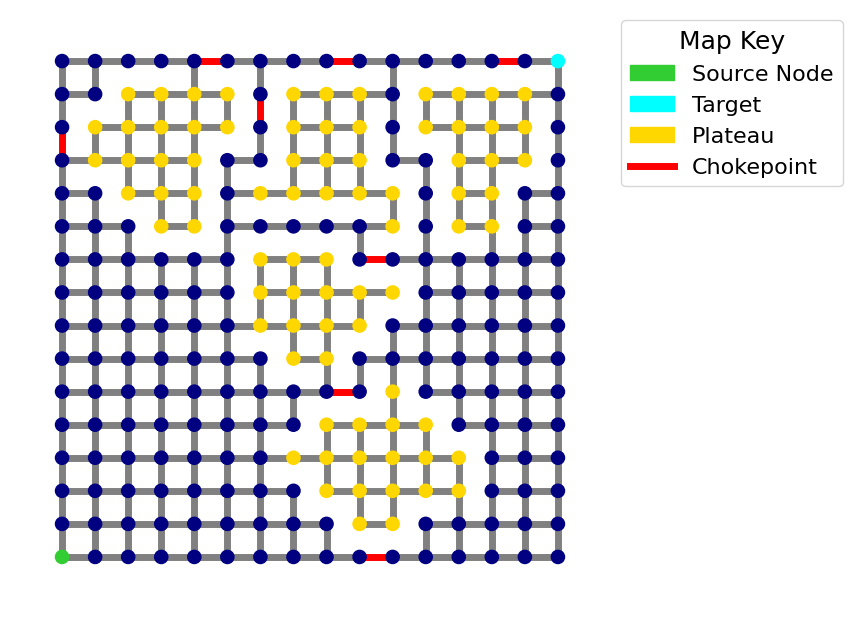}
    \caption{\textbf{Plateau Environment:} A grid environment with \textit{plateaus} (high visibility, high entry cost). Chokepoint edges have an independent blockage probability $p$ (not used during planning). }
    \label{fig:plateau_map}
    \vspace{-0.1in}
\end{figure}

We first evaluate our method on the plateau environment shown in Fig. \ref{fig:plateau_map}. This environment is motivated by the common robotics task of navigating narrow spaces between randomly placed obstacles \cite{barn_challenge}. We assume that the obstacles can be climbed at a few designated locations and refer to them as \textit{plateaus}. Climbing on top of a plateau will naturally increase visibility, but the cost of traversal will be high.

\begin{figure*}[t]
    \centering
    \includegraphics[width=\linewidth]{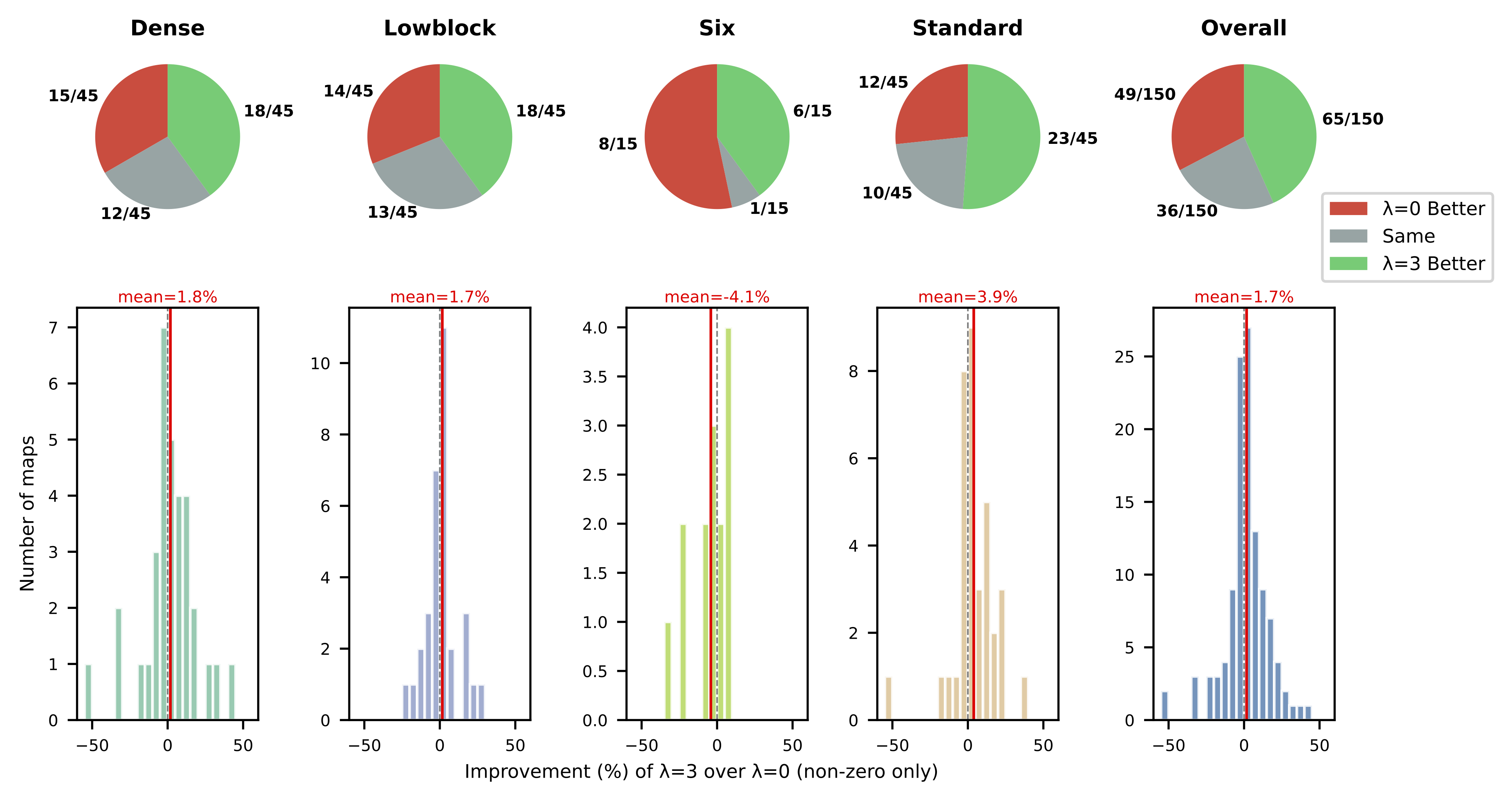}
    \caption{\textbf{Procedurally Generated Environment Test Results:} The performance improvement $(\%)$ of our method with $\lambda=3$ compared to $\lambda=0$. The top pie charts show the fraction of maps where $\lambda=3$ is better, equal or worse. The bottom histograms show the distribution of non-zero improvements.}
    \label{fig:genereated_map_histograms}
    \vspace{-0.1in}
\end{figure*}

The environment is discretized into a grid where the source node $v_s$ is located on the bottom left and the target node $v_t$ is on the opposite top right. Plateaus are placed to form narrow corridors that the agent must navigate. The plateaus can be ascended at a few specific locations; once atop a plateau, every edge within and surrounding the plateau is visible. Thus, the nodes inside these plateaus serve as high-visibility locations. All other nodes are limited to observing only the adjacent edges. To reflect the effort required to gain height, the cost to traverse from a regular node to a plateau node is 3, while the traversal cost for every other edge is 1. For simplicity, we assume symmetric edge costs. 

For edge blockages, we define a subset of \textit{chokepoint edges} that may be blocked. These critical edges are the edges with high utility $U(e)$ that are located on a corridor. Each edge is blocked with an identical and independent probability of $p=0.5$. The sampling hyperparameters for our method are set to $n = 4, r = 4$, with 1-hop virtual obstacles. 

\begin{table}[h]
    \centering
    \caption{Comparison of performance on the plateau environment of Fig. \ref{fig:plateau_map}, across 128 seeds.}
    \label{tab:singe_plateau_results}
    \begin{tabular}{ l c c c }
        \toprule
       \textbf{Agent}       & \textbf{ Mean $\mu$} & \textbf {Std Dev $\sigma$ } & \textbf{Runtime $(ms)$} \\
        \midrule
       SP                    &  43.75  & 16.05 & 2.5 \\
       RPP                   &  34.00  & 0.00  & 11,083 \\
       Ours $(\lambda = 3)$  &  36.00  & 6.00  & 94.2 \\
       \bottomrule
    \end{tabular}
\end{table}

The performance of each agent on the plateau environment is summarized in Table \ref{tab:singe_plateau_results}. Compared to the SP baseline, our algorithm with $\lambda = 3$ achieves both a lower mean cost and reduced variance. While the RPP agent demonstrates even lower mean and variance, it has a much higher runtime compared to our method. 


\subsection{Procedurally Generated Environments} \label{subsec:procedurally_gen_env}

To further validate our approach, we procedurally generate environments that mirror the graph structure of the plateau map. These maps are generated by placing and growing plateaus on the map until the space in between the plateaus forms a 1-node wide corridor. Plateaus are isolated from ground-level nodes except at two or three entry edges. Corridors are defined as sequences of intermediate nodes with exactly two intermediate neighbors, and the final 2–3 edges of each corridor are designated as chokepoints. 

We perform experiments on four variations of the plateau environment. `Standard' is the default configuration with 4 to 5 plateaus and a total plateau coverage of $40 \text{--} 45\%$. The number of chokepoints is 18 to 21 and the independent edge block probability is $p = 0.5$. The `dense' configuration increases the number of chokepoints to 21 to 24, the `lowblock' setting decreases the block probability to $p = 0.4$, and the `six' configuration increases the number of plateaus to 6. All settings except `six' are generated 45 times, 15 times each for $n=12, 14, 16$ respectively. The `six' setting is only generated at $n=16$ as the other sizes are too small to support six plateaus. Table \ref{tab:proc_genereated} details the mean and standard deviation over 200 seeds per generated map, for both agents across the procedurally generated map categories. Fig. \ref{fig:genereated_map_histograms} illustrates the distribution of performance improvements. 

\begin{table}[h]
    \centering
    \caption{Comparison of performance on procedurally generated environments}
    \label{tab:proc_genereated}
    \begin{tabular}{llcccc}
        \toprule
        \textbf{Category} & \textbf{Agent} & \textbf{Mean $\mu$} & \textbf{Std Dev $\sigma$} & \textbf{Num. Runs}\\
        \midrule
        \multirow{2}{*}{Dense}  & $\lambda = 0$ & 44.29 & 16.40 & 8653 \\
                                & $\lambda = 3$ & 43.13 & 15.62 & 8653 \\
        \midrule
        \multirow{2}{*}{Lowblock}& $\lambda = 0$ & 42.25  & 17.25 & 8718 \\
                                & $\lambda = 3$ & 41.68  & 17.94 & 8718 \\
        \midrule
        \multirow{2}{*}{Six}    & $\lambda = 0$ & 42.39  & 15.80 & 2950 \\
                                & $\lambda = 3$ & 43.91  & 16.59 & 2950 \\
        \midrule
        \multirow{2}{*}{Standard} & $\lambda = 0$ & 43.99  & 15.89 & 8681 \\
                                & $\lambda = 3$ & 42.58  & 16.59 & 8681 \\
        \midrule
        \multirow{2}{*}{\textbf{Overall}} & \boldmath$\lambda=0$   & \textbf{43.39} & \textbf{16.48} & \textbf{29002} \\
                                          & \boldmath$\lambda=3$ & \textbf{42.61} & \textbf{16.76} & \textbf{29002} \\
        \bottomrule
    \end{tabular}
\end{table}

Overall, our method with $\lambda = 3$ demonstrates lower mean cost with a slightly higher standard deviation compared to $\lambda = 0$. This trend holds for the `lowblock' configuration as well. In the `dense' configuration, $\lambda=3$ resulted in a decrease in standard deviation, which can be attributed to the increased expected number of blocked chokepoints, which makes information-gathering more beneficial. Conversely, in the `six' configuration, average performance was lower with $\lambda=3$; we hypothesize that the increased number of alternative routes reduces the cost of backtracking, thereby diminishing the relative value of preemptive observation.

\begin{figure*}[t]
    \centering
    \begin{subfigure}[b]{0.4\textwidth}
        \centering
        \includegraphics[width=\linewidth]{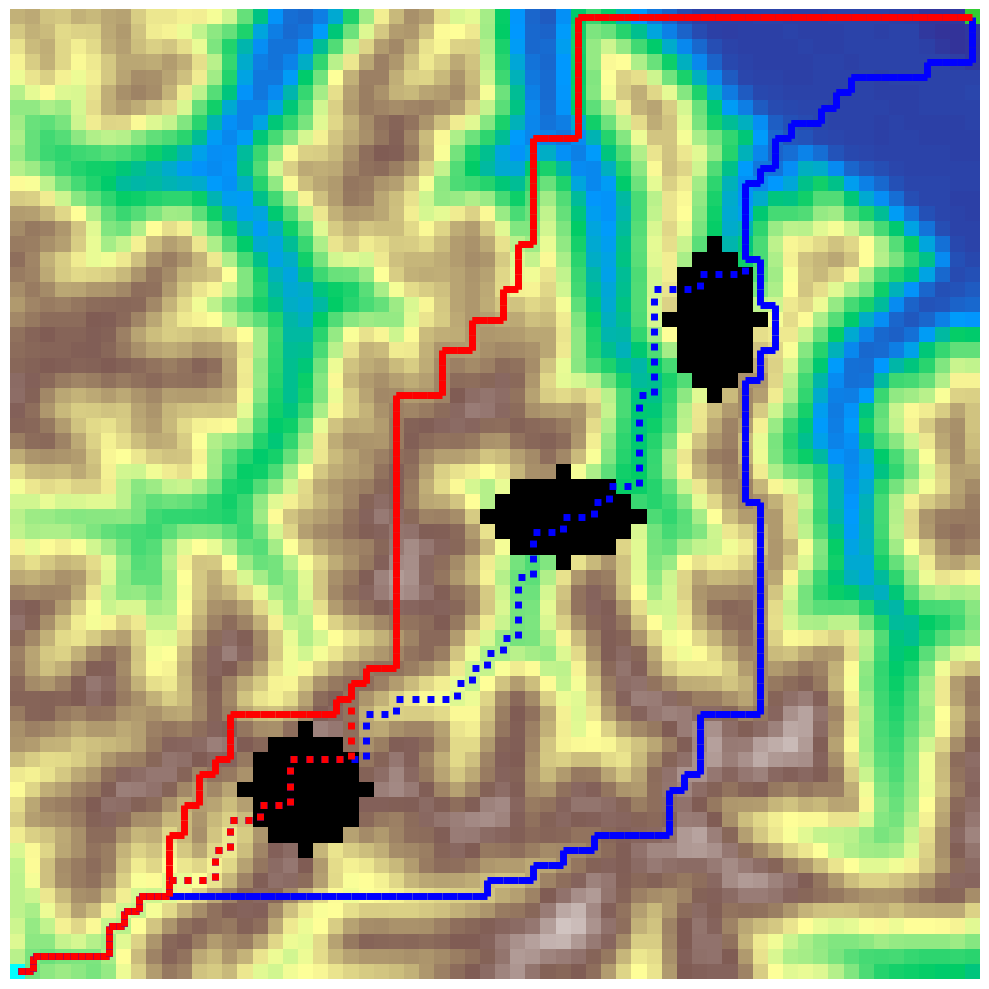}
        \label{fig:real_map_sp}
    \end{subfigure}
    \begin{subfigure}[b]{0.4\textwidth}
        \centering
        \includegraphics[width=1.15\linewidth]{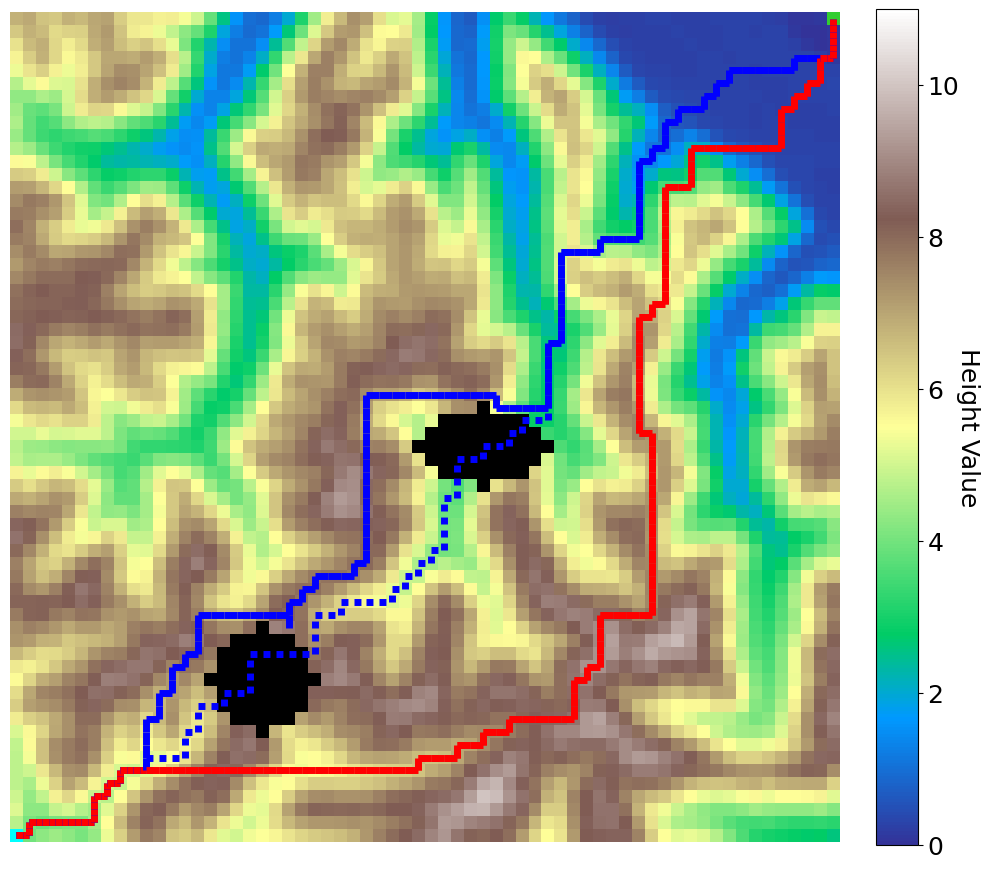}
        \label{fig:real_map_ours}
    \end{subfigure}
    \caption{\textbf{Natural Terrain Navigation:} Result of navigation from the top right to the bottom left on a real topographical map, comparing the shortest path baseline (blue) and our proposed method (red). The dashed lines show the initial intended path and the solid lines show the actual path used.}
    \label{fig:real_map_trajectory}
    \vspace{-0.1in}
\end{figure*}

Analysis of the distribution of improvement per map reveals that on most maps $\lambda = 3$ does not result in large improvement or large decrease in performance compared to $\lambda = 0$. In many instances, both reward weights yield identical results. This is expected, as not all graph structures provide high-observation vantage points that can be strategically leveraged. Overall, the results in this subsection demonstrate that our method can adapt to a variety of different maps by altering the reward weight $\lambda$.

\subsection{Natural Terrain Environment} \label{subsec:real_environment_experiment}

Next, we demonstrate the efficacy of our algorithm in more realistic environments by analyzing its performance in navigating natural terrain. Using the topography data from OpenTopography \cite{OpenTopography}, we generate a $64 \times 64$ grid graph with the node height set as the terrain height at that location. The node height is normalized to a range of $[0, 10]$. A map of Charleston, West Virginia is used as its mountainous terrain naturally results in high-visibility locations that are costly to reach. The cost of traversal of the edge $e$ from $v_1$ to $v_2$ is $c(e) = 1 + 2(h(v_2) - h(v_1))(h(v_2)-h(v_1)-0.5)$. This cost function penalizes steep ascents and sharp descents, while incentivizing slight downhill movement.

The visibility mapping is defined using the line of sight \cite{stolzle2021line_of_sight}. From node $v$, visibility in a given direction extends until it is blocked by a node of higher elevation. Thus, nodes at high elevation have a higher visibility compared to nodes at lower elevation. The objective is to navigate from $v_s$ at the top right corner to $v_t$ at the bottom left. To introduce uncertainty, three oval shaped obstacles are placed at critical locations such as the chokepoint in a valley or the center of a tall plateau. All edges within these obstacles are removed.

Fig. \ref{fig:real_map_trajectory} illustrates the trajectory of the SP baseline and our agent $(\lambda=3)$. In this scenario, all obstacles are situated along the shortest path on the unobstructed graph $G$, forcing the SP agent to perform reactive detours. The SP agent only discovers the first obstacle upon immediate encounter, leading to highly inefficient path corrections. In contrast, our agent strategically prioritizes navigating to a high-visibility vantage point early in the mission. By observing the terrain from an elevated position, our agent avoids encountering obstacles.

Table \ref{tab:real_map_results} demonstrates the mean and standard deviation of our method and the SP baseline, across different obstacle blockage probabilities $p$.

\begin{table}[ht]
    \centering
    \caption{Results ($\mu$ / $\sigma$) for varying probability of edge blockage on the natural terrain map of Fig. \ref{fig:real_map_trajectory}}
    \label{tab:real_map_results}
    \begin{tabular}{l c c c c c c }
        \toprule
        \boldmath$p$ & \multicolumn{2}{c}{\textbf{SP}} & \multicolumn{2}{c}{\boldmath$ \; \lambda = 1 \; $} & \multicolumn{2}{c}{\boldmath$ \; \lambda = 3 \; $} \\
        \cmidrule(lr){2-3} \cmidrule(lr){4-5} \cmidrule(lr){6-7} 
        & $\mu$ & $\sigma$ & $\mu$ & $\sigma$ & $\mu$ & $\sigma$ \\
        \midrule
        \boldmath$0$  & \textbf{147.62} & 0.00 & 150.49 & 1.16 & 150.80 & 1.05 \\
        \boldmath$0.2$ & 153.37 & 0.00 & \textbf{151.31} & 1.20 & 151.61 & 1.05 \\
        \boldmath$0.4$ & 157.41 & 0.00 & \textbf{152.07} & 1.34 & 152.40 & 1.00 \\
        \boldmath$0.6$ & 159.40 & 0.00 & \textbf{152.76} & 1.56 & 153.19 & 0.89 \\
        \boldmath$0.8$ & 159.01 & 0.00 & \textbf{153.38} & 1.87 & 153.97 & 0.73 \\
        \bottomrule
    \end{tabular}
\end{table}

Our agent with $\lambda > 0$ consistently outperforms the SP baseline when $p \neq 0$. The reward weights $\lambda = 1$ and $\lambda = 3$ demonstrate comparable performance: $\lambda = 1$ achieves a marginally lower mean path cost, whereas $\lambda = 3$ yields a significantly lower standard deviation. This underscores a key characteristic of our path reward: higher values of $\lambda$ place a greater priority on information gathering. While higher $\lambda$ may result in slightly longer trajectories on average, it provides a more robust policy with reduced variance by preemptively observing the uncertainties.

\subsection{Ablation Study} \label{subsec:ablation}

We investigate the sensitivity of our algorithm's behavior to the reward weight $\lambda$. For this study, we utilize a plateau environment featuring 18 chokepoints that are blocked with a probability $p$. Each configuration is evaluated across 1,000 randomized trials.

\begin{table}[ht]
    \centering
    \caption{Detailed performance breakdown ($\mu$ / $\sigma$) for various $p$ and $\lambda$}
    \label{tab:lambda_p_split}
    \begin{tabular}[width=\columnwidth]{l c c c c c c c c}
        \toprule
        \boldmath$p$& \multicolumn{2}{c}{\textbf{Shortest Path}} & \multicolumn{2}{c}{\boldmath$\; \lambda = 0 \;$} & \multicolumn{2}{c}{\boldmath$ \; \lambda = 3 \; $} & \multicolumn{2}{c}{\boldmath$ \; \lambda = 10 \; $} \\
        \cmidrule(lr){2-3} \cmidrule(lr){4-5} \cmidrule(lr){6-7} \cmidrule(lr){8-9}
        & $\mu$ & $\sigma$ & $\mu$ & $\sigma$ & $\mu$ & $\sigma$ & $\mu$ & $\sigma$ \\
        \midrule
        \boldmath$0$   & 22.00 & 0.00 & 22.00 & 0.00 & 22.00 & 0.00 & 22.00 & 0.00 \\
        \boldmath$0.1$ & 30.10 & 10.4 & 30.10 & 10.4 & \textbf{29.98} & 10.1 & 30.51 & 10.3 \\
        \boldmath$0.2$ & 35.35 & 11.7 & 35.35 & 11.7 & \textbf{34.77} & 10.5 & 34.85 & 10.4 \\
        \boldmath$0.3$ & 39.20 & 12.1 & 39.20 & 12.1 & \textbf{37.62} & 9.77 & 37.63 & 9.49 \\
        \boldmath$0.4$ & 42.46 & 11.8 & 42.46 & 11.8 & 39.71 & 8.68 & \textbf{39.63} & 8.40 \\
        \boldmath$0.5$ & 45.11 & 10.7 & 45.11 & 10.7 & 40.98 & 6.68 & \textbf{40.93} & 6.48 \\
        \bottomrule
    \end{tabular}
\end{table}

The results in Table \ref{tab:lambda_p_split} demonstrate that $\lambda = 0$ yields behavior identical to the shortest path agent. This observation is consistent with our formulation.

Table \ref{tab:lambda_p_split} further demonstrates that the optimal value of $\lambda$ depends on the probability of edge blockage. When $p$ is high, larger values of $\lambda$ improve performance by prioritizing observation over immediate cost minimization. This proactive information gathering minimizes the need for costly backtracking, ultimately reducing the total traversal cost.

While we have been comparing our algorithm against the shortest path agent, these results highlight that the shortest path baseline is actually a specific configuration of our method. By parameterizing the importance of observation through $\lambda$, our approach provides a flexible continuum between greedy navigation and exploratory behavior. This adaptability allows the system to be tuned to the specific demands of varying map topologies and blockage distributions. Furthermore, the fast inference speed of our algorithm allows for multiple preemptive simulation runs to determine the optimal $\lambda$ before deployment.



\section{LIMITATIONS}

While the use of the reward weight $\lambda$ allows for easy adjustment of the planner's prioritization of exploration, finding the optimal value remains a heuristic process involving trial and error. A method to find the optimal $\lambda$ is needed to more easily use our algorithm. Furthermore, as a sampling-based heuristic, our method carries no guarantees on suboptimality, in contrast to the competitive-ratio analyses available for classical CTP. Finally, results on the procedurally generated maps (Table \ref{tab:proc_genereated}) were mixed, with the shortest path configuration $\lambda = 0$ yielding better results on a third of the maps. The variation in improvement of $\lambda=3$ over $\lambda =0$ across different types of maps indicates that some maps have advantageous vantage points while others do not. Determining whether a given map rewards observation from map analysis alone remains an open problem.

\section{CONCLUSIONS}

In this paper, we present a fast, heuristic framework for navigating uncertain environments with heterogeneous visibility. We generate a path that balances the benefit of observation against the cost of reaching high-visibility locations by optimizing both the observation reward and the total cost of traversal. Simulation results demonstrate that our method navigates uncertain environments with a lower mean cost of traversal compared to the shortest path baseline. 

Several avenues for future work remain. Experimental results indicate that there exists an optimal $\lambda$ for a given map and an estimated probability of blockage. A fast method to tune $\lambda$ is a necessary future work for efficient implementation of our method. Additionally, our method is for a single-target, single-agent setting; an extension to multi-target, multi-agent settings is another potential future work. We hope that this work serves as a baseline for navigating uncertain environments with heterogeneous visibility.

\addtolength{\textheight}{-12cm}   




\section*{ACKNOWLEDGMENT}

This work is supported by the Office of Naval Research grant no. N-00014-25-1-2369 and Office of Naval Research grant no. N-00014-25-1-2519. We also thank Manav Vora for helping with the writing.

In accordance with the IROS guidelines on generative AI usage, we disclose that Claude (Anthropic) was used to generate portions of the simulation code used in the experiments of Section \ref{sec:experiments}, available at the linked repository, and to assist in drafting and revising the text. All AI-generated content was reviewed, edited, and verified by the authors, who take full responsibility for the final manuscript.


\bibliographystyle{IEEEtran}
\bibliography{IEEEabrv, citations}

\end{document}